\documentclass[11pt]{article}
\usepackage[preprint,hyperref]{acl}
\usepackage{times}
\usepackage{latexsym}
\usepackage{amsmath}
\usepackage{amssymb}
\usepackage{graphicx}
\usepackage{booktabs}
\usepackage{multirow}
\usepackage{xcolor}
\usepackage{subcaption}
\usepackage{algorithm}
\usepackage{algorithmicx}
\usepackage{algpseudocode}

\definecolor{posgreen}{HTML}{2E7D32}
\definecolor{negred}{HTML}{C62828}

\newcommand{\method}{LCD}
\newcommand{\methodfull}{Language-Conditional Dequantization}

\title{\methodfull{}: Recovering What Quantization \\Steals from Non-English Languages}

\author{
  Nirmal Thomas \\
  Prathama International
}

\hypersetup{
  pdfauthor={Nirmal Thomas},
  pdftitle={\methodfull{}: Recovering What Quantization Steals from Non-English Languages},
  pdfsubject={},
  pdfkeywords={}
}

\begin{document}
\maketitle

\begin{abstract}
Aggressive quantization disproportionately harms multilingual capability: in the sub-4B INT3 GPTQ regime, we measure 2--4$\times$ larger perplexity degradation on non-English languages than on English. We propose \methodfull{} (\method{}), a post-hoc method that attaches per-language rank-2 LoRA corrections to the linear layers of an already-quantized model, adding 0.12\% parameters per language and training in under 20 minutes on a single GPU. Across Qwen2.5-3B and Llama-3.2-3B, \method{} recovers 70--83\% of the perplexity gap for non-Latin-script languages and 17--28\% of the GlobalMMLU accuracy gap, outperforming a language-agnostic correction of equal capacity by 3--9 points on typologically distant languages and a data-free low-rank baseline (LQER) by an order of magnitude. We further identify a \emph{perplexity--accuracy disconnect} and trace it to \emph{where} quantization concentrates damage: early-depth errors (Llama) propagate downstream and resist local correction, while late-depth errors (Qwen) do not. A layer-restricted variant of \method{} validates this mechanism directly.
\end{abstract}

\section{Introduction}

Quantized large language models serve billions of queries daily across the world's languages. When a Korean user queries a 3-bit quantized model, they receive measurably worse outputs than an English user, not because the original model lacked Korean ability, but because the quantization process was calibrated exclusively on English data. In this paper, we quantify the severity of this gap in the regime where quantization is most needed: on Qwen2.5-3B, INT3 GPTQ degrades Arabic perplexity by 4.37$\times$ and Japanese by 3.04$\times$, compared to just 1.35$\times$ for English; on Llama-3.2-3B the pattern repeats (Arabic 3.65$\times$, English 1.39$\times$). In the sub-4B INT3 regime, this is a practically consequential, language-dependent quality gap.

Multilingual calibration \cite{chimoto2026calibrating} addresses the root cause but requires re-quantization, which is impractical for already-deployed models. Static error correction methods (LQER \cite{zhang2024lqer}, RILQ \cite{rilq2025}, ResQ \cite{resq2025}) add low-rank residuals post-hoc but apply the same correction regardless of input language. Input-conditional methods like BinaryMoS \cite{jo2024binarymos} adapt weights to the input but target binarization and ignore language identity.

We propose \methodfull{} (\method{}), built on a simple insight: \emph{if quantization error is language-dependent, the correction should be too}. For each linear layer in a quantized model, \method{} attaches a per-language rank-$r$ additive correction via forward hooks:
\begin{equation}
\hat{y} = W_q x + \frac{1}{r}(A_\ell \cdot B_\ell) \cdot x
\label{eq:lcd}
\end{equation}
where $A_\ell \in \mathbb{R}^{d_{\text{out}} \times r}$ and $B_\ell \in \mathbb{R}^{r \times d_{\text{in}}}$ are language-specific correction matrices. With rank $r=2$, this adds only 0.12\% parameters per language. Corrections are trained on 256 samples of language-specific text in under 20 minutes per language on a single GPU. At inference time, the appropriate correction is selected based on the input language, requiring no modification to the base model architecture.

Our contributions are:
\begin{enumerate}
\item We characterize the multilingual degradation caused by English-calibrated INT3 GPTQ on two sub-4B models (Qwen2.5-3B, Llama-3.2-3B), quantifying per-language perplexity ratios from 1.35$\times$ (English) to 4.37$\times$ (Arabic). While the broad phenomenon is known \cite{marchisio2024quantization}, we pin down its magnitude in the specific regime where aggressive quantization is most needed for deployment, motivating the need for language-conditional correction.
\item We propose \method{}, a post-hoc method that recovers 70--83\% of the perplexity gap for non-Latin-script languages at 0.12\% parameter cost per language, without re-quantization.
\item We show that the per-language conditioning captures real language-specific signal: a language-agnostic rank-2 baseline matches per-\method{} on average but trails by 3--9 points on the typologically distant languages where per-language corrections have the most signal to exploit.
\item We reveal a \emph{perplexity--accuracy disconnect} and trace it to a structural cause: a per-layer error analysis shows that quantization concentrates damage in different network depths across models (Qwen: late; Llama: early-middle). We \emph{validate and probe} this mechanism: \method{} restricted to Llama's bottom half of transformer blocks outperforms uniform correction by 10~pp at half the parameter cost (Table~\ref{tab:layer-strategy}); precision-targeting of the exact worst-error layers~(7--14) does not further improve over uniform, indicating that broad early-depth coverage, rather than narrow layer allocation, is the operative mechanism.\footnote{Code: \url{https://anonymous.4open.science/r/LCD-2466}}
\end{enumerate}

\section{Related Work}

\paragraph{Multilingual Quantization Harm.}
\citet{marchisio2024quantization} documented that quantization disproportionately degrades non-English performance, particularly for non-Latin-script languages.
\citet{borgersen2025english} found no such harm for k-quantized Llama-3.3-70B, but their 70B moderate-quantization regime differs qualitatively from the sub-4B INT3 setting where redundancy is scarce.
\citet{chimoto2026calibrating} addressed the root cause by using multilingual calibration data during GPTQ, but this requires re-quantization and cannot be applied to already-deployed models. \method{} is complementary: it operates post-hoc.

\paragraph{Quantization Error Correction.}
Prior work on low-rank error correction splits on a single axis: \emph{at-quantization-time} vs.\ \emph{post-hoc}.
LRQ \cite{lrq2025}, OmniQuant \cite{omniquant2024}, LQER \cite{zhang2024lqer}, QERA \cite{qera2025}, RILQ \cite{rilq2025}, and ResQ \cite{resq2025} all require access to the quantization pipeline and produce language-agnostic corrections.
Recover-LoRA \cite{recoverLora2025} is the closest prior work: it trains LoRA adapters post-hoc on a pre-quantized model to recover lost accuracy.
\method{} extends this post-hoc paradigm with language-\emph{conditional} selection: each language receives a dedicated adapter that specializes to its quantization-error profile. To our knowledge, \method{} is the only post-hoc method that is also language-conditional.

\paragraph{Input-Conditional Adaptation.}
BinaryMoS \cite{jo2024binarymos} uses token-adaptive scaling for binarized models; MLAS-LoRA \cite{mlasLora2025} applies language-aware LoRA for multilingual fine-tuning (not quantization repair). \method{} occupies the language-level position on the input-conditionality spectrum and targets a distinct problem: correcting systematic quantization error concentrated in specific language distributions.

\section{Method}

\subsection{Language-Conditional Correction}

For each linear layer and each language $\ell$, we define a rank-$r$ additive correction:
\begin{equation}
y_\text{corrected} = W_q x + \frac{1}{r} A_\ell B_\ell x
\end{equation}
where $A_\ell \in \mathbb{R}^{d_\text{out} \times r}$ and $B_\ell \in \mathbb{R}^{r \times d_\text{in}}$ are learnable parameters and $1/r$ is a fixed scaling constant. We initialize $A_\ell = \mathbf{0}$ and $B_\ell \sim \mathcal{N}(0, 0.01)$, ensuring the correction is exactly zero at initialization. This follows the LoRA parameterization \cite{hu2022lora} but serves a fundamentally different purpose: rather than adapting the model to a new task, we are \emph{correcting systematic quantization error} for a specific language distribution.

Corrections are implemented as \texttt{forward\_hook}s on each linear layer (excluding \texttt{lm\_head} and \texttt{embed\_tokens}), requiring no modification to the model architecture or forward pass.

\subsection{Training}

Each language is trained independently. Given a set of $N=256$ text samples from the target language (sourced from mC4/C4), we minimize the standard language modeling loss with only the correction parameters $\{A_\ell, B_\ell\}$ trainable:
\begin{equation}
\mathcal{L}_\ell = -\sum_{t} \log P(x_t | x_{<t}; W_q, A_\ell, B_\ell)
\end{equation}

We use AdamW with learning rate $5 \times 10^{-4}$, cosine decay with 10\% warmup, weight decay 0.01, gradient clipping at 1.0, and train for 500 steps per language. All base model parameters remain frozen. Training takes approximately 15--20 minutes per language on a single L4 GPU, so covering all eight non-English languages costs under 3 GPU-hours in total. Each trained correction ships as a $\sim$7\,MB delta on top of the \emph{unchanged} deployed checkpoint; by contrast, re-quantizing with multilingual calibration requires curating per-language calibration data, re-running the quantization pipeline, and redistributing the full multi-gigabyte model artifact.

\subsection{Inference}

At inference time, the input language is identified (via locale, automatic detection, or explicit specification) and the corresponding correction slot is activated via a single index. In many deployments the language is already known from the user locale or product surface; when it is not, off-the-shelf language identification (e.g., fastText LID) classifies a prompt in well under a millisecond, negligible relative to a single decoding step of the LLM itself. The base quantized model resides in memory once; language switching requires only updating the active adapter index, with no model reloading. For $K$ languages with rank $r=2$, total overhead is $K \times 0.12\%$ additional parameters.

\section{Experimental Setup}

\paragraph{Models.} Qwen2.5-3B \cite{qwen2.5} and Llama-3.2-3B \cite{llama3}, two sub-4B families with distinct tokenizers and training distributions.

\paragraph{Quantization.} GPTQ \cite{frantar2023gptq} W3A16, group size 128, calibrated on 128 English C4 samples, replicating the standard English-only deployment practice.

\paragraph{Languages.} 9 languages: English (baseline), Arabic, Japanese, Chinese, Hindi, Russian, French, Spanish, Korean.

\paragraph{Metrics.} (1) \emph{Perplexity} on held-out mC4/C4 (32 samples/language); degradation ratio and recovery \%. (2) \emph{GlobalMMLU accuracy} \cite{globalmmlu2024} via log-likelihood over $\approx$14{,}040 items per language (57 subjects).

\paragraph{Baselines.} FP16 (unquantized), INT3 (uncorrected), INT3+\method{} (corrected). English is the calibration language and excluded from correction training.

\section{Results}

\subsection{INT3 Quantization Disproportionately Harms Non-English}

Table~\ref{tab:diagnostic} presents the perplexity degradation ratios across languages; the underlying absolute perplexities for every condition are given in Appendix~\ref{app:raw-ppl}. On Qwen2.5-3B, English degrades by 1.35$\times$, while Arabic degrades by 4.37$\times$, Korean by 3.49$\times$, and Japanese by 3.04$\times$ (a relative disparity of 2.3--3.2$\times$). The pattern is consistent on Llama-3.2-3B, confirming that the effect is not model-specific. Languages with non-Latin scripts and greater typological distance from English exhibit the largest degradation; Romance languages (French, Spanish) suffer only modest additional harm beyond English.

\begin{table}[t]
\centering
\small
\setlength{\tabcolsep}{5pt}
\begin{tabular}{lcccc}
\toprule
& \multicolumn{2}{c}{\textbf{Qwen2.5-3B}} & \multicolumn{2}{c}{\textbf{Llama-3.2-3B}} \\
\cmidrule(lr){2-3} \cmidrule(lr){4-5}
\textbf{Lang} & I/F$\times$ & vs EN & I/F$\times$ & vs EN \\
\midrule
English  & 1.35 & --   & 1.39 & --   \\
Arabic   & 4.37 & 3.23 & 3.65 & 2.63 \\
Japanese & 3.04 & 2.25 & 2.79 & 2.01 \\
Chinese  & 2.39 & 1.77 & 2.89 & 2.07 \\
Hindi    & 2.73 & 2.02 & 2.14 & 1.54 \\
Russian  & 2.12 & 1.57 & 2.39 & 1.72 \\
French   & 1.64 & 1.21 & 1.71 & 1.23 \\
Spanish  & 1.60 & 1.19 & 1.66 & 1.20 \\
Korean   & 3.49 & 2.58 & 3.00 & 2.16 \\
\bottomrule
\end{tabular}
\caption{Perplexity degradation ratios after INT3 GPTQ quantization. I/F = INT3/FP16 ratio (mC4/C4 held-out, 32 samples/lang); ``vs EN'' is the ratio of each language's degradation to English's.}
\label{tab:diagnostic}
\end{table}

\subsection{\method{} Recovers Most of the Perplexity Gap}

Table~\ref{tab:recovery} shows the perplexity recovery results. \method{} recovers 70--83\% of the quantization gap for non-Latin-script languages (Arabic, Japanese, Chinese, Korean) on both models. Recovery is lower for Latin-script languages close to English (French 38\%/54\%, Spanish 36\%/52\%), consistent with these languages suffering less language-specific quantization error for \method{} to correct. These figures are robust to the random seed: retraining all corrections under three seeds and re-evaluating on a fixed 256-sample held-out set changes per-language recovery by at most $\pm1.3$ percentage points (standard deviation), so the results are not an artifact of a single seed.

\begin{table}[t]
\centering
\small
\setlength{\tabcolsep}{5pt}
\begin{tabular}{lccl}
\toprule
\textbf{Lang} & \textbf{Qwen} & \textbf{Llama} & \textbf{Script} \\
\midrule
Arabic   & \textcolor{posgreen}{78.8} & \textcolor{posgreen}{80.1} & Arabic \\
Japanese & \textcolor{posgreen}{69.7} & \textcolor{posgreen}{72.9} & CJK \\
Chinese  & \textcolor{posgreen}{71.2} & \textcolor{posgreen}{76.4} & CJK \\
Korean   & \textcolor{posgreen}{74.8} & \textcolor{posgreen}{76.7} & Hangul \\
Hindi    & \textcolor{posgreen}{82.6} & 68.6 & Devanagari \\
Russian  & 57.0 & \textcolor{posgreen}{71.7} & Cyrillic \\
French   & 38.2 & 53.8 & Latin \\
Spanish  & 35.6 & 52.2 & Latin \\
\midrule
\textbf{Avg (non-EN)}    & \textbf{63.5} & \textbf{69.1} & \\
\textbf{Avg (non-Latin)} & \textbf{72.4} & \textbf{74.4} & \\
\bottomrule
\end{tabular}
\caption{Perplexity recovery (\%): fraction of the INT3$\to$FP16 gap closed by \method{}. Green: $\geq$70\%. English is excluded from correction training as it is the calibration language.}
\label{tab:recovery}
\end{table}

\subsection{Language-Specific vs.\ Language-Agnostic Correction}

A natural question is whether per-language conditioning is necessary, or whether a single shared rank-2 LoRA trained on mixed multilingual data would suffice. Table~\ref{tab:agnostic} compares \method{} against a language-agnostic baseline (same rank, same training budget, training data pooled across four target languages) on Qwen2.5-3B.

\begin{table}[t]
\centering
\small
\setlength{\tabcolsep}{5pt}
\begin{tabular}{lccc}
\toprule
\textbf{Language} & \textbf{Per-lang} & \textbf{Agnostic} & \textbf{$\Delta$} \\
\midrule
Arabic   & 78.8 & 69.9 & \textcolor{posgreen}{$+$8.9} \\
Japanese & 69.7 & 62.7 & \textcolor{posgreen}{$+$7.0} \\
Korean   & 74.8 & 71.9 & \textcolor{posgreen}{$+$3.0} \\
French   & 38.2 & 54.0 & \textcolor{negred}{$-$15.8} \\
\midrule
\textbf{Average} & \textbf{65.4} & \textbf{64.6} & \textbf{$+$0.8} \\
\bottomrule
\end{tabular}
\caption{Per-language vs.\ language-agnostic rank-2 LoRA on Qwen2.5-3B (perplexity recovery \%). Averages are near-identical, but per-language wins by 3--9 points on typologically distant languages and loses 16 points on French, where no distinct per-language signal exists. $\Delta$ = per-language $-$ agnostic (percentage points).}
\label{tab:agnostic}
\end{table}

On average the two methods are nearly identical (65.4\% vs.\ 64.6\%), but the per-language breakdown is revealing: the per-language correction wins by 3--9 points on Arabic, Japanese, and Korean (precisely the languages whose distribution is most distant from the English calibration set) and loses 16 points on French. This pattern confirms that per-language conditioning captures genuine language-specific signal where such signal exists in the data; on French, where the quantization error is closer to English's, a shared correction trained on a multilingual mixture transfers better than a French-specific one estimated from 256 samples of French text alone. We interpret the near-tie on average as a sharpening, not a weakening, of the central claim: \method{} is a faithful correction of language-specific quantization error, not a free lunch that helps every language.

\paragraph{Comparison to data-free error reconstruction (LQER).}
A stronger, principled baseline is LQER \cite{zhang2024lqer}, which reconstructs the quantization error with a rank-$r$ truncated SVD of $W_\text{fp16}-W_q$ at each layer: low-rank like \method{}, but \emph{data-free}, \emph{untrained}, and language-agnostic. At the identical rank-2 budget on Qwen2.5-3B (Table~\ref{tab:lqer}), LQER recovers only 5.5\% of the perplexity gap and 13.0\% of the GlobalMMLU gap (non-English averages), versus 63.5\% and 27.8\% for \method{}; LQER is even negative on several languages (e.g.\ Russian, $-40\%$ perplexity). A static SVD of the weight error cannot capture the input-dependent, language-specific component that \method{} learns from data. This is the sharpest statement of our claim: at matched parameter budget, a \emph{trained, language-conditional} correction recovers an order of magnitude more of the perplexity gap than a \emph{data-free, agnostic} one.

\begin{table}[t]
\centering
\small
\setlength{\tabcolsep}{5pt}
\begin{tabular}{lcc}
\toprule
\textbf{Method (rank-2)} & \textbf{PPL} & \textbf{MMLU} \\
\midrule
LQER (data-free)         & 5.5 & 13.0 \\
\method{} (per-language) & \textbf{63.5} & \textbf{27.8} \\
\bottomrule
\end{tabular}
\caption{\method{} vs.\ LQER \cite{zhang2024lqer} on Qwen2.5-3B at matched rank-2 budget: non-English average recovery (\%) of the INT3$\to$FP16 gap, for perplexity (PPL) and GlobalMMLU (MMLU).}
\label{tab:lqer}
\end{table}

\subsection{Recovery Correlates with Typological Distance}

Recovery correlates with typological distance from English: non-Latin-script languages (Arabic, Japanese, Chinese, Korean) cluster at the top (70--83\%), while Latin-script languages close to English (French, Spanish) cluster at the bottom (36--54\%), with Hindi and Russian in between. This ordering is consistent across both models and confirms that \method{} corrects calibration--distribution mismatch: the further a language's activation statistics lie from English, the more concentrated the quantization error and the more \method{} has to recover.

\subsection{Rank Ablation}

We trained \method{} corrections on Qwen2.5-3B at ranks 1, 2, and 4 (Arabic and Japanese, 500 steps, $\text{lr}=5\!\times\!10^{-4}$). The average recovery across the two languages was 76.92\% (r=1), 76.98\% (r=2), and 76.84\% (r=4). Rank is not the bottleneck: the gap from rank 1 to rank 4 is within 0.1 points, indicating that the correction subspace is effectively one-dimensional on these languages. We adopt $r=2$ as a small safety margin; a rank-1 deployment would halve parameter overhead to 0.06\% per language without measurable degradation.

\subsection{Downstream Transfer: The Perplexity--Accuracy Disconnect}

Table~\ref{tab:mmlu} presents GlobalMMLU accuracy across conditions. INT3 quantization severely degrades accuracy on both models (Qwen: 64.7$\to$41.8\% English; Llama: 55.5$\to$39.9\%).

\begin{table*}[t]
\centering
\small
\setlength{\tabcolsep}{5pt}
\begin{tabular}{llccccccccc}
\toprule
\textbf{Model} & \textbf{Cond} & \textbf{EN} & \textbf{AR} & \textbf{JA} & \textbf{ZH} & \textbf{HI} & \textbf{RU} & \textbf{FR} & \textbf{ES} & \textbf{KO} \\
\midrule
\multirow{3}{*}{Qwen2.5-3B}
 & FP16
   & $64.7_{\pm0.4}$ & $48.7_{\pm0.4}$ & $52.1_{\pm0.4}$ & $59.5_{\pm0.4}$
   & $38.9_{\pm0.4}$ & $53.9_{\pm0.4}$ & $56.7_{\pm0.4}$ & $57.7_{\pm0.4}$ & $50.0_{\pm0.4}$ \\
 & INT3
   & $41.8_{\pm0.4}$ & $28.0_{\pm0.4}$ & $30.7_{\pm0.4}$ & $35.4_{\pm0.4}$
   & $26.3_{\pm0.4}$ & $29.6_{\pm0.4}$ & $33.3_{\pm0.4}$ & $35.0_{\pm0.4}$ & $28.1_{\pm0.4}$ \\
 & \method{}
   & $41.8_{\pm0.4}$
   & $31.5_{\pm0.4}^{\ddagger}$
   & $\mathbf{36.7}_{\pm0.4}^{\ddagger}$
   & $\mathbf{44.5}_{\pm0.4}^{\ddagger}$
   & $28.3_{\pm0.4}^{\ddagger}$
   & $\mathbf{35.8}_{\pm0.4}^{\ddagger}$
   & $\mathbf{40.0}_{\pm0.4}^{\ddagger}$
   & $\mathbf{41.7}_{\pm0.4}^{\ddagger}$
   & $\mathbf{35.9}_{\pm0.4}^{\ddagger}$ \\
\midrule
\multirow{3}{*}{Llama-3.2-3B}
 & FP16
   & $55.5_{\pm0.4}$ & $39.9_{\pm0.4}$ & $41.6_{\pm0.4}$ & $44.7_{\pm0.4}$
   & $38.3_{\pm0.4}$ & $44.4_{\pm0.4}$ & $48.2_{\pm0.4}$ & $48.7_{\pm0.4}$ & $41.1_{\pm0.4}$ \\
 & INT3
   & $39.9_{\pm0.4}$ & $28.5_{\pm0.4}$ & $30.4_{\pm0.4}$ & $31.7_{\pm0.4}$
   & $28.4_{\pm0.4}$ & $31.2_{\pm0.4}$ & $34.3_{\pm0.4}$ & $32.9_{\pm0.4}$ & $30.1_{\pm0.4}$ \\
 & \method{}
   & $39.9_{\pm0.4}$
   & $31.4_{\pm0.4}^{\ddagger}$
   & $31.9_{\pm0.4}^{\dagger}$
   & $32.8_{\pm0.4}^{\dagger}$
   & $30.3_{\pm0.4}^{\dagger}$
   & $34.7_{\pm0.4}^{\ddagger}$
   & $35.9_{\pm0.4}$
   & $35.7_{\pm0.4}^{\ddagger}$
   & $31.2_{\pm0.4}$ \\
\bottomrule
\end{tabular}
\caption{GlobalMMLU accuracy (\%) across nine languages. Subscripts show binomial SE ($\sqrt{p(1-p)/n}$, $n\!\approx\!14{,}040$). Significance vs.\ INT3: $^\dagger p\!<\!0.05$, $^\ddagger p\!<\!0.001$ (paired $t$-test across 57 subjects, df\,=\,56). \textbf{Bold}: \method{} gain $\geq$5 pp over INT3. Qwen closes 28\% of the FP16$\to$INT3 gap (non-EN avg); Llama closes 17\%.}
\label{tab:mmlu}
\end{table*}

On Qwen2.5-3B, \method{} delivers consistent accuracy gains across every non-English language. The largest improvements are Chinese ($+$9.1 points), Spanish ($+$6.7), Russian ($+$6.2), and Japanese ($+$6.0); Arabic, French, and Korean also gain between 3.5 and 7.8 points. Across the eight non-English languages, \method{} closes an average of 28\% of the FP16$\to$INT3 accuracy gap.

On Llama-3.2-3B, \method{} also improves every non-English language, but the gains are smaller, between 1.1 points (Chinese, Korean) and 3.5 points (Russian), closing an average of only 17\% of the gap. This \emph{perplexity--accuracy disconnect} is the paper's most scientifically informative finding: Llama's perplexity recovery (avg 69\%) exceeds Qwen's (avg 63\%), yet its MMLU recovery lags substantially. Section~\ref{sec:layerwise} diagnoses this gap using a third measurement, distinct from both perplexity and task accuracy: per-layer activation error, which reveals \emph{where} in the network quantization concentrates its damage and why that location determines whether perplexity recovery translates into downstream gains.

\subsection{Diagnosing the Disconnect: Where Quantization Hurts}
\label{sec:layerwise}

\begin{figure*}[t]
\centering
\includegraphics[width=0.95\linewidth]{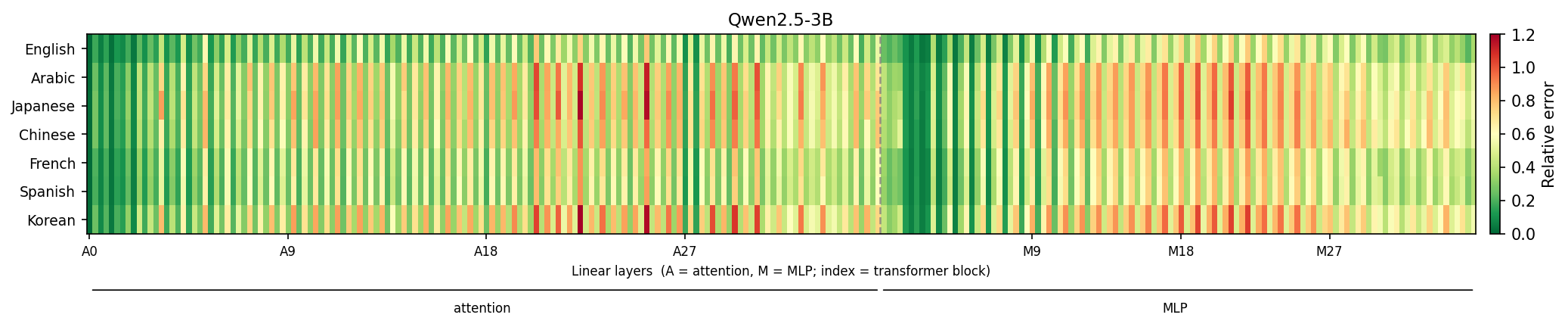}\\[3pt]
\includegraphics[width=0.95\linewidth]{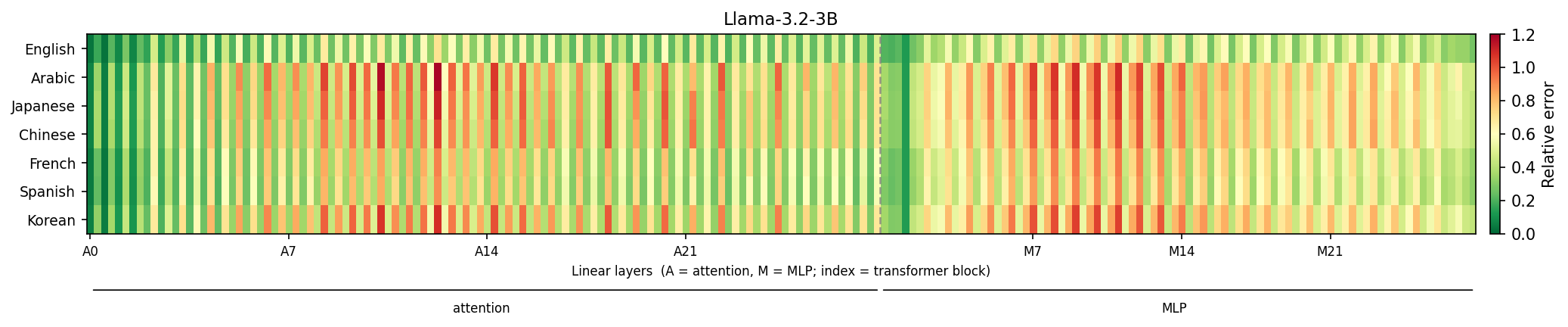}
\caption{Per-layer relative INT3 quantization error by language. Top: Qwen2.5-3B. Bottom: Llama-3.2-3B. Each column is one linear layer; columns are sorted attention-then-MLP, shallow-to-deep within each block. Color indicates $\|y_\text{fp16}-y_\text{int3}\|_F / \|y_\text{fp16}\|_F$ averaged over 32 samples per language. The worst-error layers are the ``compression'' projections (\texttt{o\_proj}, \texttt{down\_proj}) in both models, but their \emph{depth} differs: Qwen's hotspots cluster in layers 18--30 (out of 36), while Llama's cluster in layers 7--14 (out of 28).}
\label{fig:layerwise}
\end{figure*}

To understand the perplexity--accuracy disconnect mechanistically, we measure the relative Frobenius error $\|y_\text{fp16} - y_\text{int3}\|_F / \|y_\text{fp16}\|_F$ for every linear layer's output activations, separately for each language. Figure~\ref{fig:layerwise} shows the resulting heatmaps.

Two structural findings explain the disconnect. First, the highest-error layers in both models are \texttt{o\_proj} and \texttt{down\_proj}, the compression bottlenecks that project from a larger intermediate space back to the hidden dimension, where GPTQ's per-channel grid is least faithful. Second, and more consequentially, the \emph{depth} of these worst-error layers differs sharply: Qwen2.5-3B's top-10 lie in indices 18--30 of 36 (late); Llama-3.2-3B's lie in indices 7--14 of 28 (early-middle), with 14+ downstream layers consuming their corrupted output.

This depth asymmetry explains the disconnect directly. Late-layer errors (Qwen) primarily distort final next-token logits, which perplexity measures and \method{} fixes locally. Early-middle errors (Llama) distort intermediate representations that feed into downstream attention and MLP operations, exactly the multi-step computation MMLU requires. A rank-2 hook corrects a layer's immediate output but cannot undo propagated upstream corruption. Quantization damage thus operates at two levels: \emph{distributional} (perplexity, correctable regardless of depth) and \emph{representational} (downstream accuracy, correctable only when error is near the output). The representational reading also predicts a scale effect: larger models carry more redundancy, so INT3 should inflict less representational damage and leave less for \method{} to recover; our Qwen2.5-7B experiment (Limitations) bears this out, with non-English MMLU recovery falling to 8.5\% and higher-capacity corrections overfitting.

\paragraph{Direct validation: targeting the worst-error depth band.}
The depth-asymmetry hypothesis predicts that, for each model, restricting \method{}'s correction to the half of the network containing the worst-error layers should match or exceed uniform correction; restricting it to the opposite half should underperform. We test this by training \method{} (rank $r=2$, identical hyperparameters) under three layer-coverage configurations on the four languages with the largest INT3 degradation:
(i)~\textsc{all} layers (the default);
(ii)~\textsc{bottom-half} of transformer blocks (covering Llama's worst-error band at indices 7--14);
(iii)~\textsc{top-half} of transformer blocks (covering Qwen's worst-error band at indices 18--30).
Table~\ref{tab:layer-strategy} reports the resulting perplexity recovery.

\begin{table}[t]
\centering
\small
\setlength{\tabcolsep}{4pt}
\begin{tabular}{llcccc}
\toprule
\textbf{Model} & \textbf{Strategy} & \textbf{AR} & \textbf{JA} & \textbf{ZH} & \textbf{KO} \\
\midrule
\multirow{3}{*}{Qwen2.5-3B}
 & all          & 84.6 & 77.3 & 18.9 & 81.8 \\
 & bottom-half  & 74.4 & 70.2 &  3.8 & 73.1 \\
 & top-half     & 78.6 & 68.2 & $-$1.9 & 70.5 \\
\midrule
\multirow{3}{*}{Llama-3.2-3B}
 & all                       & 71.9 & 71.0 & 21.8 & 70.2 \\
 & \textbf{bottom-half}      & \textbf{78.5} & \textbf{83.3} & \textbf{36.1} & \textbf{78.3} \\
 & top-half                  & 40.2 & 45.6 & $-$20.2 & 37.6 \\
\bottomrule
\end{tabular}
\caption{Perplexity recovery (\%) for \method{} under three layer-coverage configurations, all at rank $r=2$ and identical training. \textsc{bottom-half} covers Llama's worst-error band (indices 7--14) and recovers more than uniform \textsc{all} on every Llama language despite using half the parameters. On Qwen, whose worst layers (18--30) span a larger fraction of the network, \textsc{all} dominates and the two halves are not differentiated.}
\label{tab:layer-strategy}
\end{table}

On Llama, \textsc{bottom-half} exceeds uniform \textsc{all} by 10~pp and beats \textsc{top-half} by 43~pp on average, obtaining better recovery at half the parameters. On Qwen, the two halves are indistinguishable and both trail \textsc{all}, consistent with its worst-error layers spanning a broad 40\% band where any half-subset misses some hotspots. The depth asymmetry is visible precisely because Llama's worst layers (25\% of the network) are entirely contained in one half and absent from the other.

Critically, this perplexity advantage does \emph{not} reach downstream accuracy. Evaluated on GlobalMMLU, Llama's \textsc{bottom-half} \method{} recovers only 9.5\% of the accuracy gap, versus 17\% for uniform \textsc{all} (the \emph{reverse} of the perplexity ordering). Concentrating capacity where the \emph{distributional} error is largest (the early compression band) actively trades away \emph{representational} recovery. This double dissociation (bottom-half wins on perplexity but loses on MMLU) is the strongest evidence that the two levels of quantization damage are governed by distinct mechanisms, and that perplexity recovery is an unreliable proxy for downstream gains.

\paragraph{Negative result: narrow layer targeting does not improve further.}
We follow up by testing whether concentrating corrections \emph{specifically} on layers~7--14 with additional rank yields further gains. Three configurations are evaluated: \textsc{uniform-r2} (rank-2, all layers), \textsc{targeted-r4} (rank-4, layers~7--14 only), and \textsc{targeted-r2} (rank-2, layers~7--14 only). \textsc{uniform-r2} achieves 45.0\% average recovery; both targeted variants reach only 37\%, and \textsc{targeted-r4}$\approx$\textsc{targeted-r2} (37.0\% vs.\ 37.1\%) rules out rank as an explanatory factor. The bottom-half advantage thus comes from broad coverage of the early depth band (layers~0--13 as a whole) rather than precise targeting of the identified hotspot. When correction capacity is limited, \emph{broad} early-depth coverage dominates narrow layer allocation.

\section{Conclusion}

We have shown that English-calibrated INT3 quantization creates a systematic, language-dependent quality gap that disproportionately harms non-English users. \method{} demonstrates that this gap is largely correctable: rank-2 corrections at 0.12\% parameter cost per language recover 70--83\% of the perplexity degradation for non-Latin-script languages across two model families, and 17--28\% of the GlobalMMLU accuracy gap. A language-agnostic baseline with the same capacity matches per-\method{} on average but trails by 3--9 points on the typologically distant languages where language-specific signal is concentrated: evidence that \method{} is a faithful correction of language-specific quantization error rather than a generic fine-tune.

Our analysis further reveals that perplexity recovery does not reliably translate to downstream accuracy: Llama's higher perplexity recovery yields smaller MMLU gains than Qwen's. A per-layer error analysis traces this to depth asymmetry: Llama's worst-error layers sit early-middle in the network, propagating corruption downstream, while Qwen's sit late. Restricting \method{} to Llama's bottom half confirms this (+10~pp over uniform), yet precision-targeting of layers~7--14 specifically does not help further (37\% vs.\ 45\%), establishing that broad early-depth coverage, not narrow layer allocation, is the operative mechanism.

The broader implication is practical: \emph{quantization need not discriminate}. With trivial overhead, deployed quantized models can serve non-English users more equitably.


\section*{Limitations}

\paragraph{Language identification at inference.} \method{} requires the input language to be known at inference time so that the correct per-language adapter is activated. In controlled deployment (e.g., a localized product surface with known user locale) this is straightforward, but code-switched inputs, low-quality language ID, or mixed-language prompts are not directly handled by the current design. A soft gating mechanism over adapters, or falling back to the language-agnostic adapter when confidence is low, is a natural extension but is not evaluated here.

\paragraph{Residual MMLU gap on early-error architectures.} Section~\ref{sec:layerwise} attributes \method{}'s smaller MMLU gains on Llama-3.2-3B to the depth of its worst-error layers (indices 7--14 of 28), whose corrupted outputs propagate through many downstream blocks. We evaluate a layer-targeted variant that concentrates rank-4 corrections exclusively on these layers, but it \emph{underperforms} uniform rank-2 correction (37\% vs.\ 45\% perplexity recovery), indicating that the MMLU gap reflects a structural representational deficit that narrow layer targeting cannot overcome. Allocating higher rank specifically to the compression projections (\texttt{o\_proj}, \texttt{down\_proj}), which carry the highest per-layer error in both models, rather than to entire transformer blocks, remains an open direction.

\paragraph{Narrow quantization regime.} We study INT3 GPTQ with W3A16, group size 128, English C4 calibration. Other quantization schemes (AWQ, GGUF k-quants, SmoothQuant), other bit-widths (INT2, INT4, FP4), and other calibration corpora may change both the magnitude of the language-specific harm and the degree to which a rank-2 correction suffices. We expect the qualitative story to hold but make no claim beyond the evaluated setting.

\paragraph{Model scale and family coverage.} Our main evaluation uses two sub-4B-parameter models (Qwen2.5-3B, Llama-3.2-3B). To probe scale, we additionally quantized Qwen2.5-7B to INT3 and trained \method{} corrections, evaluating on GlobalMMLU. The result is the opposite of a clean scale-up: at 7B, rank-2 \method{} recovers only 8.5\% of the non-English accuracy gap (vs.\ 28\% at 3B), and raising capacity to rank-4 with 1000 steps makes \emph{every} language worse (average $-3.9\%$), as the corrections overfit the 256-sample training slice. Crucially, the languages quantization damages most still recover at 7B (Korean $+4.7$~pts, 26.5\%; Arabic $+2.4$~pts, 17.0\%), while languages with small INT3 gaps (Russian, Spanish, French) go flat or slightly negative. We read this as consistent with, not counter to, our central claim: a larger model carries more redundancy, so INT3 inflicts less \emph{representational} damage on high-resource languages, leaving little language-specific error for a correction to recover, and additional capacity then fits noise. \method{}'s downstream benefit is therefore damage-dependent and does not automatically grow with scale; characterizing exactly where on the (scale $\times$ bit-width) plane the benefit persists is open. We make no claim about MoE architectures or instruction-tuned variants.

\paragraph{Language coverage and data.} Perplexity recovery is measured on 32 mC4/C4 held-out samples per language, and MMLU on the GlobalMMLU test set ($\approx$14{,}040 items per language). The eight non-English languages we evaluate span four scripts, but they are all relatively high-resource. Recovery behavior on truly low-resource languages, on dialectal variation, and on domain-shifted text (legal, medical, code-mixed) is an open question.

\paragraph{Adapter training data.} Each per-language adapter is trained on 256 samples of monolingual text from that language for 500 steps. This is deliberately small to match the ``post-hoc, cheap'' framing, but it also means the correction is estimated from a narrow slice of each language's distribution. Larger and more diverse per-language data may shift the per-language vs.\ agnostic comparison, particularly for languages like French where the agnostic baseline currently wins.

\paragraph{Ethical and fairness considerations.} \method{} is intended to \emph{reduce} a fairness gap introduced by English-centric calibration. However, because the correction is trained on web-scraped multilingual text, it inherits any biases and quality issues present in that corpus. We do not audit the adapters for language-specific toxicity, factuality, or stereotype changes introduced by the correction itself; such an audit is necessary before deploying \method{} in user-facing settings.

\section*{Ethics Statement}

\paragraph{Societal benefit.}
English-calibrated INT3 quantization imposes a systematic quality penalty on non-English users: perplexity degrades 2.6--3.2$\times$ more for Arabic and Korean than for English on the same model, with no corresponding reduction in model size for those users.
\method{} is designed to reduce this disparity.
Because corrections train on 256 samples in under 20 minutes on a single consumer GPU, the method is accessible to researchers without large-scale compute resources.
Because it operates post-hoc on already-quantized weights, it offers a practical correction path for deployed models where re-quantization is not an option.

\paragraph{Known limitations and risks.}
\method{} requires the input language to be identified at inference time.
Misidentification activates the wrong per-language correction; because the adapters are small (0.12\% of parameters, initialized near zero), the magnitude of harm from a mismatched adapter is bounded, but it is not zero and is not characterized in this work.
Each adapter is trained on web-scraped monolingual text (C4 or mC4).
These corpora carry societal biases, quality skew, and domain imbalance.
We do not audit whether the correction process amplifies or introduces language-specific toxicity, stereotyping, or factual error; such an audit is necessary before deploying \method{} in user-facing systems.
Our evaluation covers eight non-English languages, all relatively high-resource; generalization to endangered or very-low-resource languages is untested.
Finally, practitioners should not treat perplexity recovery as a proxy for full task recovery: on Llama-3.2-3B, 69\% average perplexity recovery translates to only 17\% MMLU gap recovery.

\paragraph{Data and compute.}
All training and evaluation data (C4, mC4, GlobalMMLU) are publicly available.
No human subjects were involved and no personally identifiable information was used.
Total compute for the experiments reported in this paper was approximately 42 GPU-hours on L4/L40S hardware.

\bibliography{references}

@inproceedings{marchisio2024quantization,
  title={How Does Quantization Affect Multilingual {LLM}s?},
  author={Marchisio, Kelly and Dash, Saurabh and Chen, Hongyu and Aumiller, Dennis and {\"U}st{\"u}n, Ahmet and Hooker, Sara and Ruder, Sebastian},
  booktitle={Findings of EMNLP},
  year={2024},
  note={arXiv:2407.03211}
}

@inproceedings{chimoto2026calibrating,
  title={Calibrating Beyond {E}nglish: Addressing Multilingual Bias in Quantization},
  author={Chimoto, Everlyn Asiko and Elhoushi, Mostafa and Bassett, Bruce},
  booktitle={Proceedings of EACL},
  year={2026}
}

@misc{borgersen2025english,
  title={English K-Quantization Does Not Disproportionately Diminish Multilingual Performance},
  author={Borgersen, Niklas and Goodwin, Morten},
  year={2025},
  howpublished={arXiv preprint arXiv:2503.03592}
}

@inproceedings{jo2024binarymos,
  title={Mixture of Scales: Memory-Efficient Token-Adaptive Binarization for Large Language Models},
  author={Jo, Dongwon and Kim, Taesu and Kim, Yulhwa and Kim, Jae-Joon},
  booktitle={NeurIPS},
  year={2024},
  note={arXiv:2406.12311}
}

@inproceedings{zhang2024lqer,
  title={{LQER}: Low-Rank Quantization Error Reconstruction for {LLM}s},
  author={Zhang, Cheng and Cheng, Jianyi and Constantinides, George A. and Zhao, Yiren},
  booktitle={ICML},
  year={2024},
  note={arXiv:2402.02446}
}

@inproceedings{rilq2025,
  title={{RILQ}: Rank-Insensitive {LoRA}-based Quantization Error Compensation for Boosting 2-bit Large Language Model Accuracy},
  author={Lee, Geonho and Lee, Janghwan and Hong, Sukjin and Kim, Minsoo and Ahn, Euijai and Chang, Du-Seong and Choi, Jungwook},
  booktitle={AAAI},
  year={2025},
  note={arXiv:2412.01129}
}

@inproceedings{resq2025,
  title={{ResQ}: Mixed-Precision Quantization of Large Language Models with Low-Rank Residuals},
  author={Saxena, Utkarsh and Sharify, Sayeh and Roy, Kaushik and Wang, Xin},
  booktitle={ICML},
  year={2025},
  note={arXiv:2412.14363}
}

@inproceedings{qera2025,
  title={{QERA}: An Analytical Framework for Quantization Error Reconstruction},
  author={Zhang, Cheng and Wong, Jeffrey T.~H. and Xiao, Can and Constantinides, George A. and Zhao, Yiren},
  booktitle={ICLR},
  year={2025},
  note={arXiv:2410.06040}
}

@inproceedings{mlasLora2025,
  title={{MLAS-LoRA}: Language-Aware Parameters Detection and {LoRA}-Based Knowledge Transfer for Multilingual Machine Translation},
  author={Dong, Tianyu and Li, Bo and Liu, Jingsong and Zhu, Shaolin and Xiong, Deyi},
  booktitle={Proceedings of ACL},
  year={2025}
}

@inproceedings{lrq2025,
  title={{LRQ}: Optimizing Post-Training Quantization for Large Language Models by Learning Low-Rank Weight-Scaling Matrices},
  author={Lee, Jung Hyun and Kim, Jeonghoon and Yang, June Yong and Kwon, Se Jung and Yang, Eunho and Yoo, Kang Min and Lee, Dongsoo},
  booktitle={Proceedings of NAACL-HLT},
  year={2025},
  note={arXiv:2407.11534}
}

@inproceedings{omniquant2024,
  title={{OmniQuant}: Omnidirectionally Calibrated Quantization for Large Language Models},
  author={Shao, Wenqi and Chen, Mengzhao and Zhang, Zhaoyang and Xu, Peng and Zhao, Lirui and Li, Zhiqian and Zhang, Kaipeng and Gao, Peng and Qiao, Yu and Luo, Ping},
  booktitle={ICLR},
  year={2024},
  note={arXiv:2308.13137}
}

@inproceedings{recoverLora2025,
  title={Recover-{LoRA}: Data-Free Accuracy Recovery of Degraded Language Models via Low-Rank Adaptation},
  author={Das, Devleena and Patwari, Rajeev and Sirasao, Ashish},
  booktitle={Proceedings of EMNLP (Industry Track)},
  year={2025},
  note={arXiv:2510.08600}
}

@article{frantar2023gptq,
  title={{GPTQ}: Accurate Post-Training Quantization for Generative Pre-trained Transformers},
  author={Frantar, Elias and Ashkboos, Saleh and Hoefler, Torsten and Alistarh, Dan},
  journal={ICLR},
  year={2023}
}

@article{hu2022lora,
  title={{LoRA}: Low-Rank Adaptation of Large Language Models},
  author={Hu, Edward J and Shen, Yelong and Wallis, Phillip and Allen-Zhu, Zeyuan and Li, Yuanzhi and Wang, Shean and Wang, Lu and Chen, Weizhu},
  journal={ICLR},
  year={2022}
}

@misc{globalmmlu2024,
  title={Global {MMLU}: Understanding and Addressing Cultural and Linguistic Biases in Multilingual Evaluation},
  author={Singh, Shivalika and Romanou, Angelika and Fourrier, Cl{\'e}mentine and others},
  year={2024},
  howpublished={arXiv preprint arXiv:2412.03304}
}

@misc{qwen2.5,
  title={Qwen2.5 Technical Report},
  author={{Qwen Team}},
  year={2024},
  howpublished={arXiv preprint arXiv:2412.15115}
}

@misc{llama3,
  title={The {L}lama 3 Herd of Models},
  author={Grattafiori, Aaron and others},
  year={2024},
  howpublished={arXiv preprint arXiv:2407.21783}
}

\appendix

\section{Absolute Perplexities}
\label{app:raw-ppl}

Tables~\ref{tab:diagnostic} and~\ref{tab:recovery} report degradation ratios and recovery percentages to normalize across languages with very different base perplexities. Table~\ref{tab:raw-ppl} provides the underlying absolute values (mC4/C4 held-out, 32 samples per language) for all three conditions.

\begin{table}[h]
\centering
\small
\setlength{\tabcolsep}{4pt}
\begin{tabular}{lcccccc}
\toprule
& \multicolumn{3}{c}{\textbf{Qwen2.5-3B}} & \multicolumn{3}{c}{\textbf{Llama-3.2-3B}} \\
\cmidrule(lr){2-4} \cmidrule(lr){5-7}
\textbf{Lang} & FP16 & INT3 & \method{} & FP16 & INT3 & \method{} \\
\midrule
English  & 13.71 & 18.56 & --    & 12.33 & 17.15 & --    \\
Arabic   &  7.99 & 34.91 & 13.70 & 11.85 & 43.31 & 18.10 \\
Japanese & 11.26 & 34.21 & 18.22 & 15.45 & 43.17 & 22.96 \\
Chinese  & 13.10 & 31.31 & 18.35 & 14.91 & 43.03 & 21.54 \\
Hindi    &  4.49 & 12.26 &  5.84 &  5.35 & 11.44 &  7.26 \\
Russian  &  5.90 & 12.51 &  8.74 &  7.74 & 18.46 & 10.77 \\
French   & 10.00 & 16.39 & 13.95 & 11.14 & 19.09 & 14.81 \\
Spanish  &  9.45 & 15.15 & 13.12 & 10.42 & 17.31 & 13.71 \\
Korean   &  7.98 & 27.85 & 12.98 & 12.92 & 38.79 & 18.94 \\
\bottomrule
\end{tabular}
\caption{Absolute perplexity per language under each condition (mC4/C4 held-out, 32 samples/language). English receives no correction, as it is the calibration language.}
\label{tab:raw-ppl}
\end{table}

\end{document}